\documentclass[letterpaper]{article} 
\usepackage[draft]{aaai2026}  
\usepackage{times}  
\usepackage{helvet}  
\usepackage{courier}  
\usepackage[hyphens]{url}  
\usepackage{graphicx} 
\usepackage{natbib}  
\usepackage{caption} 
\usepackage{amsmath}
\usepackage{algorithm}
\usepackage{algorithmic}
\usepackage{amsfonts}
\usepackage{multirow}
\usepackage{booktabs}
\usepackage{svg}
\usepackage{newfloat}
\usepackage{listings}
\DeclareCaptionStyle{ruled}{labelfont=normalfont,labelsep=colon,strut=off} 
\floatstyle{ruled}
\newfloat{listing}{tb}{lst}{}
\floatname{listing}{Listing}
\title{Unifying Locality of KANs and Feature Drift Compensation Projection for Data-free Replay based Continual Face Forgery Detection}
\author{
    Tianshuo Zhang\textsuperscript{\rm 1,2},
    Siran Peng\textsuperscript{\rm 2,1},
    Li Gao\textsuperscript{\rm 3},
    Haoyuan Zhang\textsuperscript{\rm 1,2},
    Xiangyu Zhu\textsuperscript{\rm 1,2},
    Zhen Lei\textsuperscript{\rm 1,2,4,5}\thanks{Corresponding author.}
}
\affiliations{
    \textsuperscript{\rm 1}School of Artificial Intelligence, University of Chinese Academy of Sciences\\
    \textsuperscript{\rm 2}MAIS, Institute of Automation, Chinese Academy of Sciences\\
    \textsuperscript{\rm 3}China Mobile Financial Technology Co., Ltd.\\
    \textsuperscript{\rm 4}CAIR, HKISI, Chinese Academy of Sciences\\
    \textsuperscript{\rm 5}SCSE, the Faculty of Innovation Engineering, M.U.S.T\\
    tianshuo.zhang@nlpr.ia.ac.cn, \{pengsiran2023, zhanghaoyuan2023, xiangyu.zhu, zhen.lei\}@ia.ac.cn, gaolids@chinamobile.com
}

\usepackage{bibentry}

\begin{document}

\maketitle

\begin{abstract}
The rapid advancements in face forgery techniques necessitate that detectors continuously adapt to new forgery methods, thus situating face forgery detection within a continual learning paradigm. However, when detectors learn new forgery types, their performance on previous types often degrades rapidly, a phenomenon known as catastrophic forgetting. Kolmogorov-Arnold Networks (KANs) utilize locally plastic splines as their activation functions, enabling them to learn new tasks by modifying only local regions of the functions while leaving other areas unaffected. Therefore, they are naturally suitable for addressing catastrophic forgetting. However, KANs have two significant limitations: 1) the splines are ineffective for modeling high-dimensional images, while alternative activation functions that are suitable for images lack the essential property of locality; 2) in continual learning, when features from different domains overlap, the mapping of different domains to distinct curve regions always collapses due to repeated modifications of the same regions. In this paper, we propose a \textbf{KAN}-based \textbf{C}ontinual Face \textbf{F}orgery \textbf{D}etection (KAN-CFD) framework, which includes a Domain-Group KAN Detector (DG-KD) and a data-free replay Feature Separation strategy via KAN Drift Compensation Projection (FS-KDCP). DG-KD enables KANs to fit high-dimensional image inputs while preserving locality and local plasticity. FS-KDCP avoids the overlap of the KAN input spaces without using data from prior tasks. Experimental results demonstrate that the proposed method achieves superior performance while notably reducing forgetting.
\end{abstract}

\section{Introduction}
The rapid development of AI-generated content (AIGC) accelerates the evolution of face manipulation methods, making it challenging for the public to discern forged media and raising significant societal concerns. Static models~\cite{uia,sbi,seeable,laanet,yanvideo} trained on fixed datasets exhibit limited generalization capabilities, rendering them ineffective against these rapidly evolving forgery techniques. Continual learning methods~\cite{der,dgr,ewc} enable a model to be continuously updated over a sequence of tasks, representing a key strategy for detection models to keep pace with the evolution of forgery methods. However, when learning new tasks, these models often suffer a severe performance drop on previous tasks, a phenomenon known as catastrophic forgetting. Most continual face forgery detection models~\cite{dfil} mitigate this issue by retaining and replaying a small set of data from prior tasks, but this approach introduces additional privacy risks.

\begin{figure}[t]
    \centering
    \small
    \includegraphics[width=0.46\textwidth]{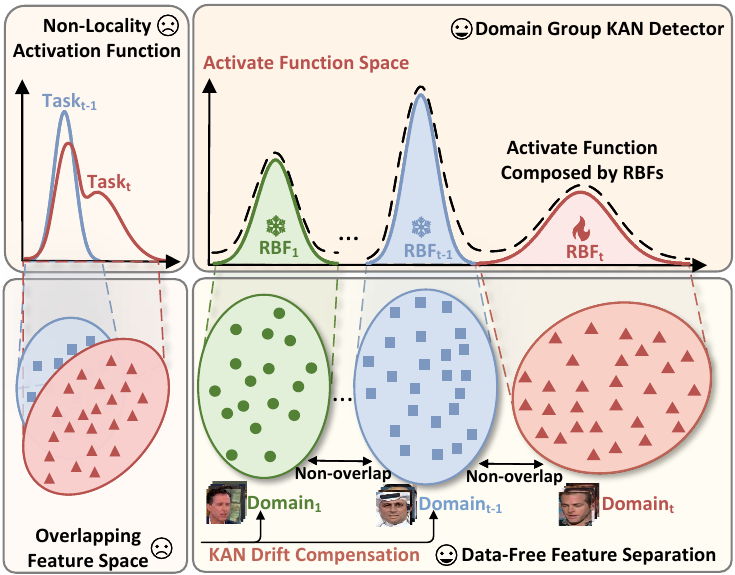}
    \caption{\small Feature space overlap and the non-local activation functions in image-KANs hinder the application of KANs in continual learning. FS-KDCP separates the feature distributions of different tasks without requiring data replay. DG-KD models the mapping from each input dimension to each output dimension for a given domain as an independent RBF. These RBFs are then combined to form a final activation function that exhibits locality, ensuring that the learning for different tasks do not interfere with each other.}
    \label{fig:first}
    \vspace{-6pt}
\end{figure}

Kolmogorov-Arnold Networks (KANs)~\cite{kan} replace the linear weights and global activation functions of MLPs~\cite{mlp} with learnable activation functions located on the edges of the network, demonstrating superior performance compared to MLPs across a wide range of applications~\cite{kan_cont2,convkan}. Specifically, the learnable activation functions in KANs exhibit locality and local plasticity. In sequential tasks, this property enables the network to adjust local curves to acquire knowledge of new tasks while preserving stability in other regions, making KANs a naturally suitable architecture for continual learning.

However, two primary limitations impede the application of KANs to continual learning scenarios. First, the original B-spline activation functions are computationally expensive, and their basis functions are ill-suited for modeling the high-dimensional data distributions characteristic of vision tasks. Several existing methods offer efficient alternatives for this activation function, such as radial basis functions (RBFs)~\cite{rbf} and rational functions~\cite{rkan}; however, they all lack the property of locality that is essential for continual learning~\cite{kac}. Second,~\cite{wisekan} demonstrates that in sequential tasks, overlapping feature distributions from different tasks result in repeated modifications of the same spline regions, which leads to the erasure of knowledge acquired from previous tasks. This issue becomes particularly severe in domain-incremental settings like face forgery detection, where the classification objective causes the feature distributions of different domains to significantly overlap.

In this paper, we introduce a KAN-based framework for Continual Face Forgery Detection (KAN-CFD) to address the aforementioned limitations. As illustrated in Fig.~\ref{fig:first}, we first propose a novel Domain-Group KAN Detector (DG-KD) that utilizes a combination of domain-specific RBFs to form DG-Layers, leveraging RBFs to enable the fitting of high-dimensional image inputs. By combining locally non-zero RBFs from different regions, it provides the essential properties of locality and plasticity required for continual learning. Second, we introduce a data-free replay based Feature Separation strategy via KAN Drift Compensation Projection (FS-KDCP) to ensure a non-overlapping input space for DG-KD. FS-KDCP employs a KAN projection (KDCP) to model the feature drift of the backbone and transform stored features into the current feature space to compensate for this drift. These transformed features are then used to estimate the overall distributions of each previous domain. Finally, the strategy separates the feature distributions of the new and old tasks to yield a non-overlapping feature space. This approach allows different domains to be modeled in distinct regions of DG-KD without requiring data from prior tasks. In summary, our main contributions are:

\begin{itemize}
    \item We propose the Domain-Group KAN Detector (DG-KD), which leverages a combination of domain-specific RBFs to model high-dimensional image inputs while providing the essential locality and local plasticity.
    \item We introduce a data-free replay based Feature Separation strategy via KAN Drift Compensation Projection (FS-KDCP), creating a non-overlapping input space for the DG-KD without requiring data from prior tasks.
    \item Extensive experiments demonstrate that our proposed framework establishes a new state-of-the-art (SOTA), achieving superior detection accuracy and the lowest forgetting rate compared to all previous methods.
\end{itemize}

\section{Related Works}
\subsection{Kolmogorov-Arnold Networks}
Kolmogorov-Arnold Networks (KANs)~\cite{kan} demonstrate superior performance compared to traditional MLPs across a wide range of tasks. However, the original KAN architecture is not well suited for vision applications due to challenges in modeling high-dimensional data. To enhance efficiency, subsequent research, including rKAN~\cite{rkan} and FastKAN~\cite{fastkan}, investigates the substitution of B-splines with rational basis functions, while KAT~\cite{kat} introduces the GroupKAN architecture, which utilizes shared activation functions across different dimensions. In the field of continual learning, WiseKAN~\cite{wisekan} reveals that feature overlap in the KAN input space leads to catastrophic forgetting. KAC~\cite{kac} presents a KAN classifier that achieves high performance in Class-Incremental Learning (CIL).

\subsection{Continual Face Forgery Detection}
Continual face forgery detection is typically formulated as a domain-incremental learning problem. Prevailing approaches primarily rely on knowledge distillation and data replay. For instance, CoReD~\cite{cored} introduces a distillation loss to preserve task-specific knowledge. DFIL~\cite{dfil} proposes a novel data selection strategy for replay and achieves competitive performance by employing distillation on both feature and label representations. DMP~\cite{dmp} introduces prototype learning and a prototype-guided replay strategy to preserve knowledge of past tasks. HDP~\cite{hdp} utilizes Universal Adversarial Perturbations to replay fake samples, while SUR-LID~\cite{surlid} achieves high performance by employing a separation and alignment strategy.

\section{Methodology}
\begin{figure*}[t]
    \centering
    \includegraphics[width=1\textwidth]{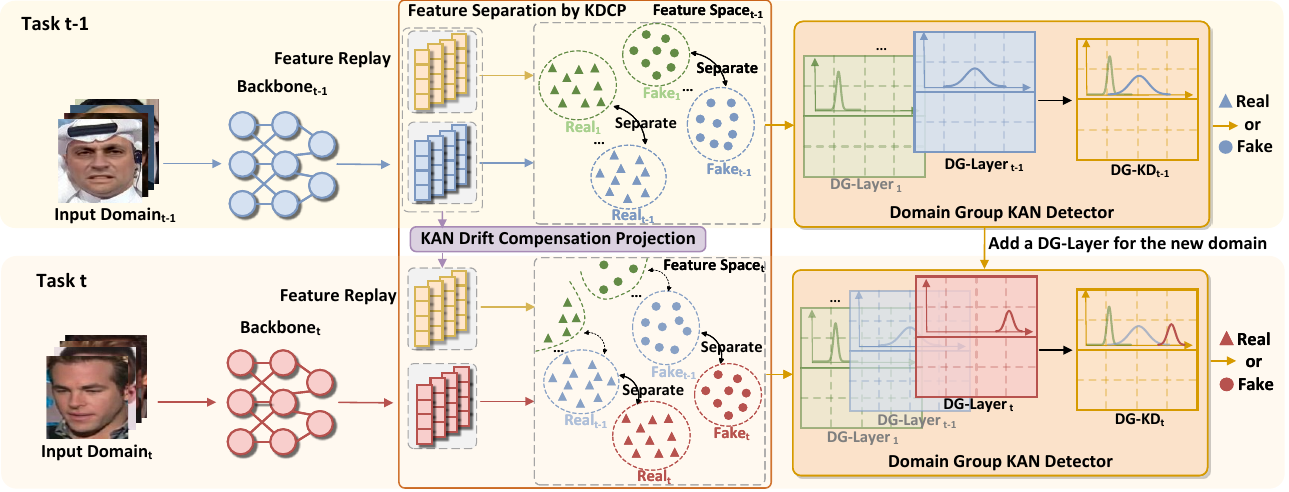}
    \caption{\small 
    Our KAN-CFD architecture retain the features of previous tasks from the feature space of $\textit{Task}_{t-1}$ when performing the $\textit{Task}_{t}$. To mitigate feature drift from an evolving backbone, we propose a data-free replay Feature Separation strategy by KAN Drift Compensation Projection (FS-KDCP). This approach employs a KAN projection to map stored features into the current feature space, which compensates for the drift and subsequently separates the feature distributions of new and old tasks. The separated features are then fed into a novel Domain-Group KAN Detector (DG-KD). The detector utilizes a separate DG-Layer to model the new task, which is then combined with the layers from previous tasks. The feature separation and the locality of the DG-KD ensure that tasks are modeled without interfering with each other.}
    \label{fig:mian}
\end{figure*}
\subsection{Preliminaries}
\subsubsection{Continual Face Forgery Detection.}
We present forgery detection as a problem of domain incremental learning, denoted as $\mathcal{D}=\left\{\mathcal{D}_{1}, \ldots, \mathcal{D}_{T}\right\}$, with $T$ representing the number of tasks. For each task $\mathcal{D}_{t}=\left\{\left(\boldsymbol{x}_{i, t}, y_{i, t}\right)\right\}_{i=1}^{n_{t}} = \{\mathcal{X}_t,\mathcal{Y}_t\}$, $\boldsymbol{x}_{i, t} \in \mathcal{X}_t$ represents the samples, and $y_{i,t}\in \mathcal{Y}_t$ denotes their labels, with $n_{t}$ being the number of examples in the task. The model processes each task $\mathcal{D}_{t}$ to learn how to handle novel forms of forgery. In the context of continual face forgery detection, tasks are characterized by varying data distributions but consistently utilize two labels:
\begin{equation}
        p\left(\mathcal{X}_{i}\right) \neq p\left(\mathcal{X}_{j}\right) \text { for }\  \mathcal{Y}_{i}=\mathcal{Y}_{j}=\text{Real or Fake} \text{ and } i \neq j,\\
    \label{equa:blending_fomular}
\end{equation}
where $p\left(\mathcal{X}_{i} \right)$ denotes the sample distribution of task $\mathcal{D}_{i}$.
\subsubsection{KAN Preliminaries.}
The Kolmogorov-Arnold Representation Theorem~\cite{kanrep} posits that any multivariate continuous function $f(\boldsymbol{x})$ defined on a bounded domain can be represented as a finite composition of univariate continuous functions combined through addition:
\begin{equation}
     f(\boldsymbol x)= f({x}_1, {x}_2, ..., {x}_n) 
     = \sum\limits_{q=1}^{2n+1}\Phi_q \Big(\sum\limits^{n}_{p=1}\phi_{q,p}(x_p)\Big),
\end{equation}
in which $\Phi_q$ and $\phi_{q,p}$ are univariate functions corresponding to each variable. The KAN~\cite{kan} architecture introduces the concept of a KAN layer. For an input vector with dimensions $d_{in}$, a KAN layer is defined as a matrix of learnable activation functions with an input dimension of $d_{\text{in}}$ and an output dimension of $d_{\text{out}}$:
\begin{equation}
    \begin{array}{ll}
         &   \text{KAN}(\boldsymbol{x}) =  \begin{bmatrix}\sum_{i=1}^{d_{in}} \phi_{1,i}(x_i) & \dots& \sum_{i=1}^{d_{in}} \phi_{d_{out},i}(x_i)\end{bmatrix} \\
         &\\
         & =\Phi \circ \boldsymbol{x} ,\text{where} \quad \Phi = \begin{bmatrix}
\phi_{1,1}(\cdot) & \cdots & \phi_{1,d_{\text{in}}}(\cdot) \\
\vdots & \ddots & \vdots \\
\phi_{d_{\text{out}},1}(\cdot) & \cdots & \phi_{d_{\text{out}},d_{\text{in}}}(\cdot)
\end{bmatrix}.
    \end{array}
    \label{eq:kan_layer}
\end{equation}
Here, $\boldsymbol{x}$ represents the inputs to the KAN layer, while $\Phi$ denotes the one-dimensional univariate function matrix of the KAN layer. The authors parameterize each univariate function using a B-spline~\cite{bspline}: 
\begin{equation}
\phi(x) = w_b \text{silu}(x) + w_s \sum_i\omega_iB_i(x).
\end{equation}
Since spline bases are local, samples $\boldsymbol{x}_{i, t} \in \mathcal{D}_{t}$ in a feature space where domains $\{\mathcal{D}_{i}\}_{i=1}^{T}$ are well-separated influence only a few adjacent spline coefficients, without affecting distant coefficients that contain prior knowledge. However, the sparse and recursive computations inherent to B-splines complicate their parallelization on GPUs, posing a challenge for processing high-dimensional inputs. Therefore, we first design a KAN architecture that is capable of processing image data while preserving the necessary locality for domain-incremental learning. Second, we construct a feature space in which the feature distributions of different tasks are distinctly separated without requiring data replay.

\subsection{Architecture Overview}
Our architecture comprises two main components: the Domain-Group KAN Detector (DG-KD) and the data-free
replay Feature Separation strategy by KAN Drift Compensation Projection (FS-KDCP). As illustrated in Fig.~\ref{fig:mian}, DG-KD sequentially models information from each domain $\mathcal{D}_{t}=\left\{\left(\mathbf{x}_{i, t}, y_{i, t}\right)\right\}_{i=1}^{n_{t}}$ by assigning a unique DG-Layer to the learning domain and subsequently combining these layers. A DG-Layer consists of an RBF matrix with group-wise parameter sharing. If the feature spaces for each domain are sufficiently distinct, the activation function, formed by a combination of locally non-zero RBFs, exhibits locality and local plasticity. The FS-KDCP module is designed to enforce separation between the feature spaces of new and old tasks. We only store extracted features from previous tasks, making the method applicable to scenarios where the original data is inaccessible. However, as the backbone network evolves, these stored legacy features are susceptible to semantic drift and may no longer accurately represent their original distributions. To address this issue, we introduce a KAN projection (KDCP) module trained on the new task. This projection aligns the feature spaces of the current and previous backbones and then maps the stored legacy features into the current feature space to reconstruct their original data distributions. A separation loss is then applied to distinguish between the feature distributions of the new and old tasks. Finally, samples from the new task are processed by DG-KD to facilitate domain modeling and classification.

\subsection{Domain Group KAN Detector}
We propose the Domain-Group KAN detector, an architecture designed for domain-incremental learning. It preserves the essential locality of B-splines while being trainable on image data. To construct activation functions with this desired locality, we select RBFs, as they are inherently local, with a non-zero response only within a finite region. A KAN layer constructed using RBFs can be formulated as:
\begin{equation}
    \begin{array}{ll}
         &   f(\boldsymbol{x})= \begin{bmatrix}\sum_{i=1}^{d_{in}} \phi_{1i}({x}_i) & \dots& \sum_{i=1}^{d_{in}} \phi_{d_{out}i}({x}_i)\end{bmatrix},\\
         &\\
         & \text{where}\quad \phi_{ij}({x}_i) = \mathrm{exp} \Big(-\frac{{({x}_i - c_{ij})}^2}{2\sigma_{ij}^2}\Big).
    \end{array}
    \label{eq:kan_rbf}
\end{equation}
Building on the premise that RBFs are easily optimized, KAT~\cite{kat} identified computational redundancy in KANs, which assign an independent learnable activation function to each input dimension. Consequently, they proposed partitioning the dimensions into $g$ groups, with each group sharing the same learnable activation function to process its $d_g = \lfloor \text{dim}(x)/g \rfloor$ dimensions. Our work addresses a critical problem of KANs: their representational capacity is often redundant in the feature dimension while being insufficient to model key domain information. Therefore, we first introduce the Domain Group Layer (DG-Layer),which is composed of a $d_{in} \times d_{out}$ matrix of RBFs. The matrix is partitioned into $g$ groups, where all RBFs within each group share the same weights:
\begin{equation}
        \begin{array}{ll}
        & \text{DG-Layer}_t(\mathbf{x})\\
        &= W \times \begin{bmatrix}  \phi_{\lfloor1/d_g\rfloor,t}(x_1) & \dots& \phi_{\lfloor d_{in}/d_g\rfloor,t}(x_{d_{in}}) \end{bmatrix}^\top ,\\
        &\\
        & where \quad W = \begin{bmatrix}
\text{w}_{1,1} & \cdots & \text{w}_{1,d_{\text{in}}} \\
\vdots & \ddots & \vdots \\
\text{w}_{d_{\text{out}},1} & \cdots & \text{w}_{d_{\text{out}},d_{\text{in}}} 
\end{bmatrix},
\end{array}
\end{equation}
where $\phi_{\lfloor i/d_g \rfloor, t}()$ represents the shared RBF for the $\lfloor i/d_g \rfloor$-th dimension group within domain/task $t$. We define the Domain Group KAN Detector (DG-KD) as a weighted sum of DG-Layers, with each layer designed to capture the information corresponding to a particular domain:
\begin{equation}
\begin{array}{ll}
        & \text{DG-KD}(\mathbf{x})\\
        & = \sum_{k=1}^{t} \text{DG-Layer}_k(\mathbf{x})\\
        &= W \times \begin{bmatrix}  \Phi_{\lfloor i/d_g\rfloor}(x_1) & \dots& \Phi_{\lfloor i/d_g\rfloor}(x_{d_{in}}) \end{bmatrix}^\top ,\\
        & where \quad \Phi_{\lfloor i/d_g\rfloor}(x_i) = \sum_{k=1}^{t} w_{{\lfloor i/d_g\rfloor},k} \cdot \phi_{{\lfloor i/d_g\rfloor},k}({x}_i),
\end{array}
\end{equation}
\begin{figure}[t]
    \centering
    \includegraphics[width=0.46\textwidth]{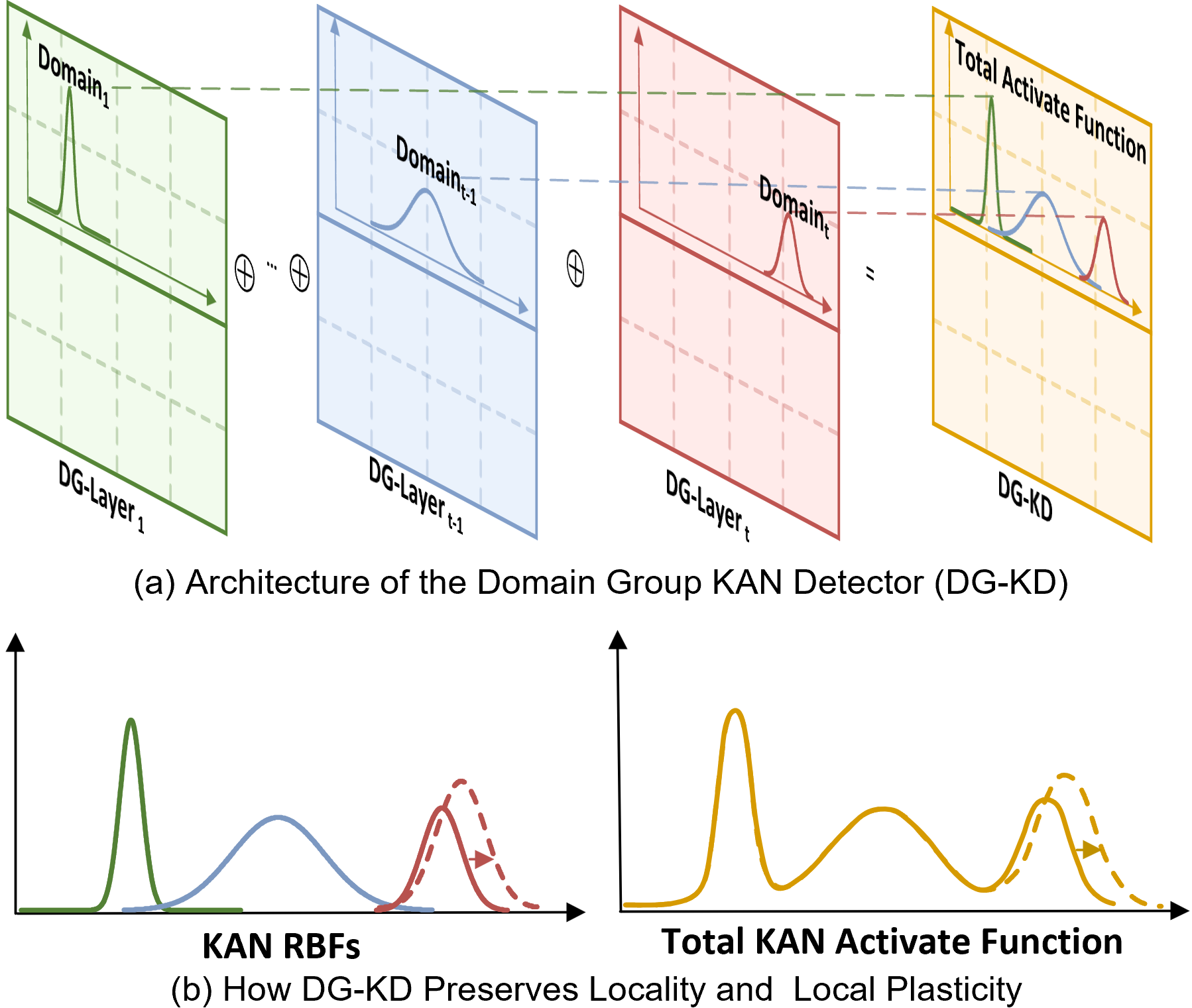}
    \caption{\small Our DG-KD (a) assigns an independent matrix of RBFs, which we term a DG-Layer, to each separated domain. All DG-Layers are then combined to form the final DG-KD. Under the assumption of separated feature spaces, (b) learning a new task (domain) only results in a local restructuring of the total activation function and does not affect its global structure.}
    \label{fig:kan_layer}
\end{figure}
We illustrate our DG-KD with an example where the feature dimensions are $d_{in}=4$, $d_{out}=4$, and the number of groups is $g=2$. As presented in Fig.~\ref{fig:kan_layer}(a), we construct a corresponding stack of DG-Layers for \( t \) tasks. For each output dimension, each layer contains $g=2$ RBFs, and each RBF processes $d_{in}/g = 2$ feature dimensions. The final output is formed by a linear combination of RBFs from corresponding positions across all layers. Under the assumption that the input spaces for each task are well-separated, this structure achieves locality and local plasticity. As depicted in Fig.~\ref{fig:kan_layer}(b), the left panel shows the example RBFs for each task, while the right panel shows the composite output activation function. Learning a new task results in only a local modification to this final function. In essence, we use a linear combination of RBFs to create an efficient implementation of B-spline-like locality.
\subsection{Feature Separation by KAN Drift Compensation Projection}
DG-KD successfully preserves locality and local plasticity while accommodating image-based inputs. However, it is still necessary to construct a well-structured feature space in which the feature distributions of different domains are non-overlapping. Such separation ensures that the curves within the DG-KD are not repeatedly modified in the same regions, thereby preventing catastrophic forgetting. Existing methods~\shortcite{si,er} typically enforce this feature separation through data replay. Our approach forgoes storing raw data and instead retains only a compact set of representative features from past tasks, which broadens its applicability. After the model completes training on $\mathcal{D}_{t-1}$, we freeze the parameters of this model to serve as a teacher. We then apply the SUR~\shortcite{surlid} method to select a sparse and robust set of representative features for $\mathcal{D}_{t-1}$:
\begin{equation}
   \{\mathbf{f}_{t-1 \rightarrow t-1}\} = \text{SUR}(f_{\theta}^{t-1}(\mathcal{X}_{
t-1})),
\end{equation}
where $\{\mathbf{f}_{t-1 \rightarrow t-1}\}$ denotes the selected features extracted from data samples $\mathcal{X}_{
t-1}$ in the feature space of $\mathcal{D}_{t-1}$, and $f_{\theta}^{t-1}$ serves as the backbone of the model for $\mathcal{D}_{t-1}$. When learning on task $\mathcal{D}_{t}$, the stored static features gradually deviate from their original positions. Prior research~\cite{drift,ldc,adc} often formulates this issue as a feature semantic drift problem. To address this problem, we employ a projection module $p_{KAN}^{t}$ that consists of a single DG-Layer to compensate for the drift and map $\{\mathbf{f}_{t-1 \rightarrow t-1}\}$ into the current feature space:
\begin{equation}
   \{\mathbf{f}_{t-1 \rightarrow t}\}=p_{KAN}^{t}(\{\mathbf{f}_{t-1 \rightarrow t-1}\}).
\end{equation}
Here, $\{\mathbf{f}_{t-1 \rightarrow t-1}\}$ denotes the selected features extracted from data samples $\boldsymbol{x}_{i, t-1}$ in the feature space of $\mathcal{D}_{t}$. We then use feature augmentation~\cite{surlid} to reconstruct the original distribution. We utilize a supervised contrastive loss to separate features from the previous and current tasks:
\begin{equation}
   \mathcal{L}_{SC}=-\frac{1}{N} \sum_{i=1}^{N} \log \left(\frac{\exp
\left(\mathbf{f}_{i} \cdot \mathbf{f}_{j} / \tau\right)}{\sum_{k=1}^{N}
\mathbb{I}_{\left[d_{i} \neq d_{k}\right]} \exp \left(\mathbf{f}_{i} \cdot
\mathbf{f}_{k} / \tau\right)}\right),
\end{equation}
where $d_i$ and $d_k$ are the domain labels. We assign unique labels to the real and fake samples corresponding to each of the $T$ tasks, resulting in a total of $2T$ labels. The terms $\mathbf{f}_i$ and $\mathbf{f}_j$ represent features that share the same domain label.

Additionally, to constrain the backbone from changing significantly, thereby enabling the projection module to accurately fit its evolution, we introduce a feature-level knowledge distillation loss. This loss utilizes data from the new task to regularize the backbone:
\begin{equation}
   \mathcal{L}_{KD}=\frac{1}{N}
\sum_{i=1}^{N}\text{MSE}\left(f_{\theta}^{t-1}\left(x_{i}\right), f_{\theta}^{t}\left(x_{i}\right)\right),
\end{equation}
where $N$ denotes the batch size, and MSE represents the mean squared error loss. Finally, we apply a binary cross-entropy loss $\mathcal{L}_{CLS}$ for classification. The overall loss for our main architecture is as follows:
\begin{equation}
   \mathcal{L}_{\text {Overall}}= \mathcal{L}_{CLS}+\lambda_{1}
\mathcal{L}_{SC}+\lambda_{2}\mathcal{L}_{KD},
\end{equation}
KDCP is trained on new task data using an alignment loss:
\begin{equation}
   \mathcal{L}_{Align}=\frac{1}{N}
\sum_{i=1}^{N}\text{MSE}\left(p_{KAN}^{t}\left(f_{\theta}^{t-1}\left(x_{i}\right)\right),f_{\theta}^{t}\left(x_{i}\right)\right).
\end{equation}
This loss minimizes the MSE between the features extracted by the current backbone $f_{\theta}^{t}$ and the mapped features from the frozen backbone of the previous task $f_{\theta}^{t-1}$, thereby aligning the feature spaces of the two tasks.

\section{Experiments}
\subsection{Datasets} We choose a variety of datasets containing forged faces from multiple domains. FaceForensics++ (FF++)~\cite{ffpp} includes 1,000 real videos. These videos are manipulated using four forgery methods: DF, F2F~\shortcite{f2f}, FS, and NT~\shortcite{NT}. The Deepfake Detection (DFD)~\cite{dfd} dataset provides over 1,000 real and forged face videos. The Deepfake Detection Challenge Preview (DFDC-P)~\cite{dfdc} contains 5,000 videos utilizing two forgery techniques. Celeb-DF v2 (CDF2)~\cite{celeb} comprises 590 genuine and 5,639 forged videos. DF40~\cite{df40} is an extensive dataset featuring 40 distinct forgery methods. Based on these datasets, we establish two evaluation protocols: a dataset incremental protocol and a forgery-type incremental protocol to assess our model's performance. For the dataset incremental protocol, we follow the benchmark proposed by DFIL and select the sequence [FF++, DFDC-P, DFD, CDF2]. For the forgery-type incremental protocol, we strictly adhere to the SUR-LID and select three forgery techniques from DF40: Face-Swapping (FS) with BlendFace, Face-Reenactment (FR) with MCNet, and Entire Face Synthesis (EFS) with StyleGAN3, along with the hybrid forgeries from FF++, collectively constitute the [Hybrid, FR, FS, EFS] testing sequence.

\subsection{Implementation Details}
For data preparation, we utilize standard datasets and data preprocessing methods from DeepFakeBench~\cite{deepfakebench} to ensure a fair comparison with existing approaches. For feature extraction, we employ ConvNeXt-B~\cite{convnext} as the backbone. For training, we use the Adam~\cite{adam} optimizer as our main optimizer, with parameters set to $\beta_1 = 0.9$ and $\beta_2 = 0.999$. The learning rate for the main optimizer is set to 2e-4. For the projection optimizer, we use the same Adam optimizer but with a learning rate of 5e-4 to promote faster convergence. The size of our feature memory is 500, which is consistent with existing methods. The batch size is set to 64. The loss hyperparameters are set to $\lambda_1 = 2$, $\lambda_2 = 1$, and $\tau = 0.1$. All experiments are conducted on two A6000 Ada GPUs.

\subsection{Metrics and Comparison Fairness}
We evaluate overall performance using Accuracy (Acc) and Area Under the Curve (AUC). The key evaluation metric, Average Forgetting (AF) quantifies the model's ability to retain information from previously learned $T$ tasks. It is defined as: $A F=\frac{1}{T} \sum_{i=1}^{T}\left(Score_{i}^{\text {first }}-Score_{i}^{\text {last }}\right)$, where $Score$ denotes either Acc or AUC. The backbone is an integral part of each method, making it infeasible to enforce uniformity. However, for the compared methods, we report the best results from officially published papers, and the key metric AF is relatively insensitive to the backbone. Therefore, we contend that our comparison is both valid and fair.

\subsection{Comparison Experiments with SOTA methods}
First, we conduct experiments under the dataset-incremental protocol, comparing our method against a comprehensive set of methods using accuracy (Acc) as the evaluation metric. The compared methods include general continual learning methods such as LWF~\cite{lwf}, as well as continual face forgery detection methods like CoReD~\cite{cored}, DFIL~\cite{dfil}, DMP~\cite{dmp}, and SUR-LID~\cite{surlid}. All continual face forgery detection methods employ data replay strategies. In contrast, our method retains only features, reducing the cache size by 99.48\%. The primary experimental results are presented in Table~\ref{tab:cross_dataset}. By the end of the training sequence, continual face forgery detection methods show considerable forgetting; CoReD and DFIL register average forgetting rates of 11.42\% and 7.01\%, respectively. Our method demonstrates superior performance, achieving the highest average Acc of 91.64\% and the lowest AF of 4.08\%. 

\begin{table}
\setlength{\tabcolsep}{0.75mm}
\small
\renewcommand\arraystretch{1.07}
\centering
\begin{tabular}{c|c|cccccc} 
\hline
\multirow{2}{*}{Method} & \multirow{2}{*}{Dataset} & \multicolumn{4}{c}{Acc(\%)$\uparrow$} & \multirow{2}{*}{Avg$\uparrow$} & \multirow{2}{*}{AF$\downarrow$}\\ \cline{3-6} 
     &  & FF++  & DFDCP & DFD   & CDF2    \\
\hline
     \multirow{4}{*}{\shortstack{$\text{LWF}^*$\\ TPAMI'\\\shortcite{lwf}}}       & FF++   & 95.52     & -      & -     & -   & 95.52     & -     \\ 
\cline{2-2}
        & DFDCP  & 87.83     & 81.57      & -     & -    & 84.70     & 7.69       \\ 
\cline{2-2}
        & DFD     & 76.16     & 41.78      & 96.36     & -    & 71.43     & 19.89      \\ 
\cline{2-2}
             & CDF2    & 67.34     & 67.43      & 84.05     & 87.90    & 76.68     & 14.44      \\ 
\hline
      \multirow{4}{*}{\shortstack{CoReD\\ MM'\\\shortcite{cored}}}       & FF++    & 95.50 & -      & -     & -     & 95.50 & -      \\ 
\cline{2-2}
       & DFDCP  & 92.94 & 87.61  & -     & -     & 90.28 & 2.56      \\ 
\cline{2-2}
  & DFD     & 86.84 & 81.07  & 95.22 & -     & 87.71 & 7.60     \\ 
\cline{2-2}
             & CDF2    & 74.08 & 76.59  & 93.41 & 80.78  & 81.22 & 11.42      \\
\hline

     \multirow{4}{*}{\shortstack{DFIL\\ MM'\\\shortcite{dfil}}}          & FF++    & 95.67 & -      & -     & -     & 95.67 & -      \\ 
\cline{2-2}
             & DFDCP  & 93.15 & 88.87  & -     & -     & 91.01 & \underline{2.52}      \\ 
\cline{2-2}
  & DFD     & 90.30 & 85.42  & 94.67 & -     & 90.03 & 4.41      \\ 
\cline{2-2}
             & CDF2    & 86.28 & 79.53  & 92.36 & 83.81  & 85.49 & 7.01      \\
\hline
     \multirow{4}{*}{\shortstack{DMP\\ MM'\\\shortcite{dmp}}}         & FF++    & 95.96 & -      & -     & -     &95.96 & -      \\ 
\cline{2-2}
             & DFDCP  & 92.71 & 89.72  & -     & -     & 91.22 & 3.25      \\ 
\cline{2-2}
   & DFD     & 92.64 & 86.09  & 94.84 & -     & 91.19 & \underline{3.48}      \\ 
\cline{2-2}
             & CDF2    & 91.61 & 84.86  & 91.81 & 91.67  & 89.99 & \textbf{4.08}      \\
\hline

\multirow{4}{*}{\shortstack{SUR-LID\\ CVPR'\\\shortcite{surlid}}} & FF++    & 96.52 & -      & -     & -     & \underline{96.52} & -      \\ 
\cline{2-2}
         & DFDCP  & 93.35 & 90.70  & -  & -  &  \underline{92.03} & 3.17      \\ 
\cline{2-2}
& DFD     & 91.58 & 87.25  & 97.50 & -   & \underline{92.11} & 4.20      \\ 
\cline{2-2}
     & CDF2    & 90.50 & 88.08  & 92.96 & 92.91 & \underline{91.11} & 4.39     \\
\hline
 \multirow{4}{*}{\shortstack{\textbf{\textit{$\text{KAN-CFD}^{*}$}}\\\textbf{\textit{(Ours)}}}}& FF++    & 97.68 & -      & -     & -     & \textbf{97.68} & -      \\ 
\cline{2-2}
        & DFDCP  & 95.90 & 90.39  & -     & -   &  \textbf{93.15} & \textbf{1.78}     \\ 
\cline{2-2}
 & DFD   & 93.52 & 88.83 & 97.69 & -     & \textbf{93.68} &  \textbf{2.86}     \\ 
\cline{2-2}
  & CDF2    & 92.90 & 87.36  & 93.26 &  93.03 & \textbf{91.64} & \textbf{4.08}    \\
\hline
\end{tabular}
\caption{\small Experiments on dataset incremental protocol. The best performer is highlighted in boldface, while the second-best result is underlined. *denotes data-free replay based methods.}
\label{tab:cross_dataset}
\vspace{-3pt}
\end{table}
\begin{table}
\setlength{\tabcolsep}{0.9mm}
\renewcommand\arraystretch{1.07}
\small
\centering
\begin{tabular}{c|c|cccccc} 
\hline
\multirow{2}{*}{Method} & \multirow{2}{*}{Type} & \multicolumn{4}{c}{AUC(\%)$\uparrow$} & \multirow{2}{*}{Avg$\uparrow$} & \multirow{2}{*}{AF$\downarrow$}\\ \cline{3-6} 
     &  & Hybrid  & FR & FS   & EFS    \\
\hline
   \multirow{4}{*}{\shortstack{$\text{LWF}^*$\\ TPAMI'\\\shortcite{lwf}}}          & Hybrid    & 97.00     & -      & -     & -     & \underline{97.00}    & -      \\ 
\cline{2-2}
            & FR  & 88.76     & 88.45      & -     & -     & \underline{88.61}     & 8.24       \\ 
\cline{2-2}
      & FS     & 84.07     & 80.99      & 96.44     & -     & 87.24     & 10.20      \\ 
\cline{2-2}
             & EFS    & 78.73     & 56.73      & 93.67     & 92.82     & 79.32     & 17.59      \\ 
\hline
  \multirow{4}{*}{\shortstack{CoReD\\ MM'\\\shortcite{cored}}}           & Hybrid    & 96.65     & -      & -     & -     & 96.65    & -      \\ 
\cline{2-2}
          & FR  & 93.55     &79.88      & -     & -     & \underline{88.61}     & \underline{3.10}      \\ 
\cline{2-2}
     & FS     & 89.07     & 79.29      & 86.05     & -     & 84.80     & 4.09      \\ 
\cline{2-2}
             & EFS    & 84.54     & 64.29      & 84.17     & 92.63     & 81.41     & 9.86      \\ 
\hline
    \multirow{4}{*}{\shortstack{DFIL\\ MM'\\\shortcite{dfil}}}         & Hybrid    & 96.46     & -      & -     & -     &96.46   & -      \\ 
\cline{2-2}
       & FR  & 55.74     & 99.75      & -     & -     & 77.75     & 40.72       \\ 
\cline{2-2}
     & FS     &60.71     & 66.49      & 99.03     & -     & 75.41     & 34.51      \\ 
\cline{2-2}
             & EFS    & 50.83     & 95.56      & 70.81     & 99.96     & 79.29     & 26.01      \\ 
\hline
   \multirow{4}{*}{\shortstack{HDP\\ IJCV'\\\shortcite{hdp}}}          & Hybrid    & 96.71     & -      & -     & -     & 96.71    & -      \\ 
\cline{2-2}
           & FR  & 67.41     &95.45      & -     & -     & 81.43     & 29.30       \\ 
\cline{2-2}
    & FS     & 63.00     & 71.35     & 95.09     & -     & 76.48     & 28.91      \\ 
\cline{2-2}
             & EFS    & 59.89     & 70.06      & 89.34     & 93.73     & 78.26     & 22.65      \\ 
\hline
     \multirow{4}{*}{\shortstack{SUR-LID\\ CVPR'\\\shortcite{surlid}}}        & Hybrid    & 96.85     & -      & -     & -     & 96.85    & -      \\ 
\cline{2-2}
            & FR  & 82.91     & 92.42      & -     & -     & 87.66     & 13.94       \\ 
\cline{2-2}
     & FS     &  90.50   & 96.26     & 97.94     & -     & \textbf{94.90}     & \textbf{1.26}      \\ 
\cline{2-2}
             & EFS    & 87.90     & 96.79      & 93.56     & 99.07     & \underline{94.33}     & \underline{2.99}      \\ 
\hline
   \multirow{4}{*}{\shortstack{\textbf{\textit{$\text{KAN-CFD}^{*}$}}\\\textbf{\textit{(Ours)}}}}          & Hybrid    & 97.02     & -      & -     & -     & \textbf{97.02}    & -      \\ 
\cline{2-2}
       & FR  & 93.37     & 92.42     & -     & -     &\textbf{92.90}     & \textbf{3.65}       \\ 
\cline{2-2}
     & FS     & 90.86     & 91.23     & 96.43     & -     & \underline{92.84}     & \underline{3.68}     \\ 
\cline{2-2}
             & EFS    & 90.78     & 91.81     & 95.47    & 99.53    & \textbf{94.40}     & \textbf{2.60}      \\ 
\hline
\end{tabular}
\caption{\small Experiments on forgery-type incremental protocol. The best performer is highlighted in boldface, while the second-best result is underlined. *denotes data-free replay based methods.}
\label{tab:cross_type}
\end{table}
Second, we perform comparative experiments under the forgery-type incremental protocol [Hybrid, FR, FS, EFS]. We include another continual forgery detection method, HDP~\cite{hdp} for comparison, using Area Under the Curve (AUC) as the evaluation metric. The experimental results are detailed in Table~\ref{tab:cross_type}. Several compared methods exhibit a severe performance decline early in training. For example, the average forgetting of DFIL after the second task is 40.72\%, while SUR-LID and HDP display AFs of 13.94\% and 29.30\%, respectively. In contrast, our method maintains a consistently low forgetting rate, concluding with a final AF of only 2.60\% and achieving the highest AUC of 94.40\%. The results from both the dataset-incremental and forgery-type incremental protocols indicate that our method achieves SOTA, surpassing even replay-based methods.

\subsection{Long-sequence Continual Learning Experiment}
To evaluate the capability of our model on long-sequence tasks, we conduct experiments with 10 tasks. These tasks are constructed from different forgery methods in DF40~\shortcite{df40}, covering three major forgery categories: FR, FS, and EFS. For comparison, we select a recent continual forgery detection method, DFIL~\shortcite{dfil}, and a recent KAN-based continual learning method, KAC~\shortcite{kac}. The detailed setup is provided in the supplementary materials. As shown in Fig.~\ref{fig:long}, our approach achieves the highest average accuracy and the lowest average forgetting rate compared to both methods, demonstrating its effectiveness in long-sequence tasks.
\begin{figure}[t]
  \vspace{-5pt}
  \centering
   \includegraphics[width=0.95\linewidth]{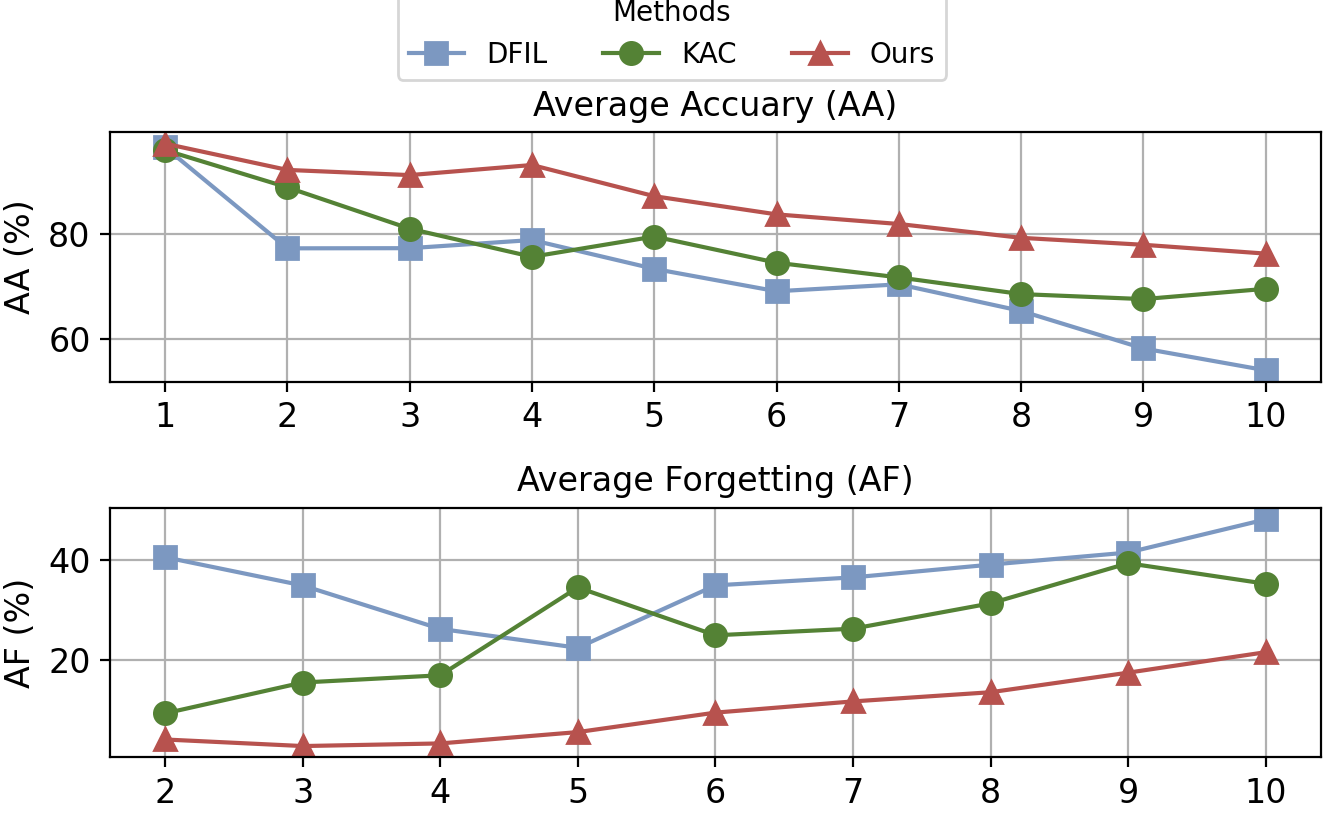}
   \caption{\small The long-sequence continual learning experiment, the proposed method achieves the highest average accuracy and the lowest forgetting rate.}
   \label{fig:long}
   \vspace{-5pt}
\end{figure}
\subsection{Ablation Study}
\subsubsection{Ablation Study on the Effect of Loss Functions.} We keep the overall framework unchanged and perform the ablation by setting the hyperparameters of each loss term to zero. In Table~\ref{tab:abla}, we find that a model trained using only the classification loss is highly susceptible to catastrophic forgetting, leading to a high AF of 31.35\% at the end of the training sequence. Both the $\mathcal L_{ {sc}}$, which serves to separate different domains, and the $\mathcal L_{ {kd}}$, designed to preserve knowledge from previous tasks within the backbone, effectively mitigate forgetting. Through the combination of these losses, our complete loss function ultimately achieves the best results. 
\begin{table}
\centering
\setlength{\tabcolsep}{0.9mm}
\renewcommand\arraystretch{1.07}
\small
\begin{tabular}{c|c|cc|cc|cc} 
\hline
\multirow{2}{*}{Loss} & FF++ & \multicolumn{2}{c|}{DFDC-P}     & \multicolumn{2}{c|}{DFD}     & \multicolumn{2}{c}{CDF2}      \\ 
\cline{2-8}
 & AA                    & AA             & AF            & AA             & AF            & AA             & AF             \\ 
\hline
$\mathcal L_{CLS}$ & 97.63                & 88.97          & 12.00          & 80.82          & 20.43         & 70.51          & 31.35          \\ 
\hline
$\mathcal L_{CLS}+\mathcal L_{KD}$ & 97.04                 & 90.85          & 5.41        & 89.33          & 7.60        & 85.56         & 13.44          \\ 
\hline
$\mathcal L_{CLS}+\mathcal L_{SC}$ & 96.37               & 91.94          & 4.26         & 89.69          & 8.01         & 87.87          & 10.64          \\ 
\hline
 $\mathcal L_{ {Overall}}$  & \textbf{97.68}      & \textbf{93.15} & \textbf{1.78} & \textbf{93.68} & \textbf{2.86} & \textbf{91.64} & \textbf{4.08 } \\
\hline
\end{tabular}
\vspace{-2pt}
\caption{\small Ablation study on the effect of loss functions.}
\label{tab:abla}
\end{table}
\begin{figure*}[t]
    \centering
    \vspace{-6pt}
    \includegraphics[width=1\textwidth]{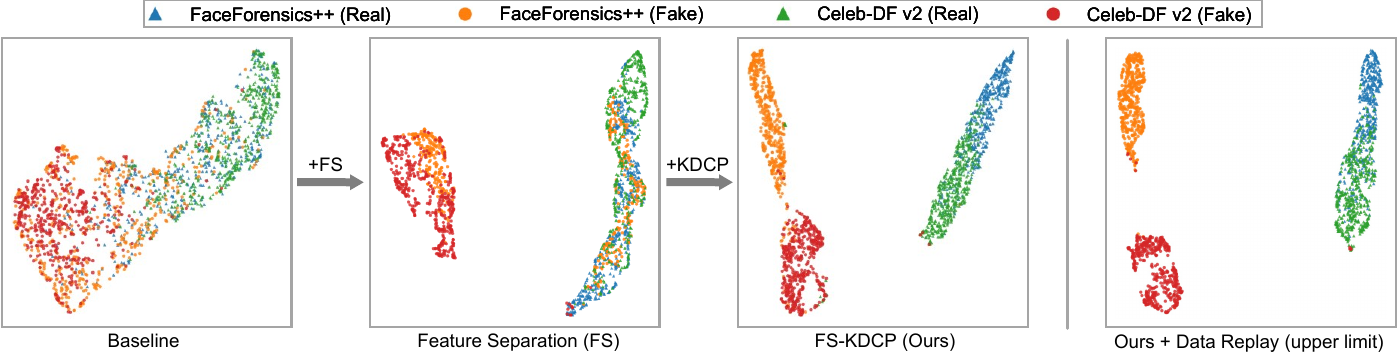}
    \caption{\small A visual comparison of Feature Separation. The first three plots from the left illustrate the progressive changes in the feature space as Feature Separation (FS) and KAN Drift Compensation Projection (KDCP) are added. The final plot represents the performance upper limit. It is observable that our method nearly preserves the feature space structure of the approach that uses data replay.}
    \label{fig:visualization}
\end{figure*}
\subsubsection{Ablation and Visualization of the Feature Separation.}
\label{exp:fi}
\begin{table}
\small
\centering
\setlength{\tabcolsep}{0.9mm}
\renewcommand\arraystretch{1.07}
\begin{tabular}{c|c|cc|cc|cc} 
\hline
 \multirow{2}{*}{Module} & FF++ & \multicolumn{2}{c|}{DFDC-P}     & \multicolumn{2}{c|}{DFD}     & \multicolumn{2}{c}{CDF2}      \\ 
\cline{2-8}
 & AA                    & AA             & AF            & AA             & AF            & AA             & AF             \\ \hline
 Baseline & 97.63                & 88.97          & 12.00          & 80.82          & 20.43         & 70.51          & 31.35          \\ 
\hline
 FS w/o KDCP & 97.42    & 87.82  & 12.46        & 84.99          & 14.80        & 83.64   & 16.04         \\
 \hline
 Ours& \textbf{97.68}    & \textbf{93.15} & \textbf{1.78} & \textbf{93.68} & \textbf{2.86} & \textbf{91.64} & \textbf{4.08 } \\
 \hline
 Ours+Replay & 97.36      & 93.37 & 1.07 & 93.65 & 2.15 & 92.63 & 3.49  \\
\hline
\end{tabular}
\caption{\small Ablation study on the feature space constraint.}
\vspace{-7pt}
\label{tab:abla2}
\end{table}
Second, we keep the DG-KD unchanged and conduct ablation and visualization experiments to analyze the separation of our feature space. The Baseline represents the backbone combined with the DG-KD, without any constraints on the feature space. 'FS w/o KDCP' entails feature separation but omits drift compensation. 'Ours+Replay' denotes the overall framework but stores raw data from previous tasks, serving as the performance upper bound. As shown in Table~\ref{tab:abla2}, the baseline model suffers from severe catastrophic forgetting. Furthermore, while incorporating feature separation helps to isolate domains, the lack of drift compensation allows the stored features to drift over time, leading to similar performance degradation. After introducing KDCP, our method closely approximates the performance upper bound.

To further validate whether KAN-CFD can effectively separate the feature spaces of different tasks, we conduct the visualization experiment shown in Fig.~\ref{fig:visualization}. The model sequentially learns two tasks, FF++ and CDF2, and we use UMAP~\cite{umap} for dimensionality reduction and visualization. We observe that the baseline model suffers from catastrophic forgetting on the FF++ task, resulting in significant feature overlap. Although using only 'Feature Separation (FS)' maintains a relatively clear classification boundary between the two datasets, the FF++ features undergo drift, causing their distribution to significantly overlap with the CDF2 features. This overlap results in some fake samples crossing the decision boundary, leading to poor performance on FF++. In contrast, our method, which incorporates the KAN projection to compensate for drift, effectively preserves the feature space structure of the upper bound without requiring the storage of raw data.

\subsubsection{Analysis of the Domain Group KAN Detector.}
Finally, we maintain the FS-KDCP and evaluate the performance of different detectors on continual face forgery detection. The results are presented in Table~\ref{tab:abla3}. A standard MLP~\shortcite{mlp} employs global activation functions and therefore lacks locality. Consequently, its performance on previous tasks degrades sharply. The traditional KAN~\cite{kan} is inapplicable because its B-splines are unsuitable for fitting high-dimensional data. Moreover, GroupKAN~\cite{kat}, which uses rational functions as its activation functions, also suffers from catastrophic forgetting due to its loss of locality. KAC~\cite{kac}, a method designed for class-incremental learning, achieves satisfactory results due to its robustness to domain-incremental settings. In contrast, our DG-KD, specifically designed for the continual forgery detection task, surpasses KAC and achieves the best results.
\begin{table}
\centering
\small
\setlength{\tabcolsep}{0.9mm}
\renewcommand\arraystretch{1.07}
\begin{tabular}{c|c|cc|cc|cc} 
\hline
 \multirow{2}{*}{Module} & FF++ & \multicolumn{2}{c|}{DFDC-P}     & \multicolumn{2}{c|}{DFD}     & \multicolumn{2}{c}{CDF2}      \\ 
\cline{2-8}
 & AA                    & AA             & AF            & AA             & AF            & AA             & AF             \\ 
\hline
MLP & 96.48      & 88.38 & 8.74 & 78.91 & 22.76 & 72.36 & 28.29  \\
\hline
GroupKAN & 97.16      & 88.91 & 8.33 & 82.20 & 17.56 & 74.42 & 25.00  \\\hline
KAC & 97.29      &91.54 & 4.21 & 91.57 & 4.45 & 86.58 & 9.77  \\\hline
DG-KD(Ours) & \textbf{97.68}    & \textbf{93.15} & \textbf{1.78} & \textbf{93.68} & \textbf{2.86} & \textbf{91.64} & \textbf{4.08 } \\
\hline
\end{tabular}
\caption{\small Analysis of the Domain Group KAN Detector.}
\vspace{-7pt}
\label{tab:abla3}
\end{table}

\section{Conclusion}
In this paper, we propose KAN-CFD, a novel KAN-based method for continual face forgery detection. Our approach overcomes the limitations of KANs, efficiently implementing the locality of KAN and adapting it for domain-incremental learning scenarios. First, we employ KAN drift compensation projection to map old features into the new feature space, thereby achieving feature space separation without data replay. Second, building upon the separated feature space, we introduce the Domain Group KAN Detector. This detector retains the crucial locality of KAN while enabling the training in image data. Extensive experiments demonstrate that our method achieves SOTA performance.

\bibliography{aaai2026}

\end{document}